\documentclass[12pt]{article}

\usepackage[authoryear, sort&compress, round]{natbib}
\usepackage{graphicx}
\usepackage[small]{caption}

\usepackage[hyphens]{url}
\usepackage[hidelinks]{hyperref}
\usepackage[hyphenbreaks]{breakurl}

\usepackage{geometry}

\usepackage{titlesec}
\usepackage{authblk}

\titleformat{\section}{\large\bfseries}{\thesection}{1em}{}

\title{Still ``Talking About Large Language Models'': Some Clarifications}

\author[1,2]{Murray Shanahan \thanks{m.shanahan@imperial.ac.uk}}
\affil[1]{Imperial College London}
\affil[2]{Institute of Philosophy, School of Advanced Study, University of London}

\date{November 2024}

\begin{document}

\maketitle

\begin{abstract}
My paper {\em Talking About Large Language Models} has more than once been interpreted as advocating a reductionist stance towards large language models. But the paper was not intended that way, and I do not endorse such positions. This short note situates the paper in the context of a larger philosophical project that is concerned with the (mis)use of words rather than metaphysics, in the spirit of Wittgenstein's later writing.
\end{abstract}

In \citep{shanahan2024talking}, I wrote ``[a] bare-bones LLM does not really know anything because all it does, at a fundamental level, is sequence prediction''. Looking at that sentence in isolation, a reader might be forgiven for assuming that I am taking some sort of reductionist stance according to which an LLM-based chatbot, such as ChatGPT, Claude, or Gemini, is {\em just} a next token predictor, where the word ``just'' here carries great metaphysical weight, and that LLM-based systems therefore do not and cannot have beliefs.\footnote{See Downes et al. (\citeyear{downes2024llms}).} There are several other sentences in that paper of a similar kind, and with hindsight, I wish I had taken greater care not to express myself in ways that are so easily open to misreading. So I am grateful for the opportunity, here, to set the record straight.\footnote{Another example of such a sentence is ``[A] great many tasks that demand intelligence in humans can be reduced to next-token prediction with a sufficiently performant model'' \citep[p.68]{shanahan2024talking}. It would have been better to have written ``cast as next-token prediction'' rather than ``reduced to next-token prediction'', and not to have expressed the point in a way that seems to preclude describing LLMs as ``intelligent''.}

First, and most importantly, I would like to make explicit the overarching philosophical project I see myself as engaged in, which is very much in the spirit of Wittgenstein's later work, exemplified by the {\it Philosophical Investigations} \citep{wittgenstein1953philosophical}. Generally speaking, I dislike all philosophical claims of the form X is Y (or X is not Y) where the word ``is'' carries metaphysical weight. In my 2010 book, {\em Embodiment and the Inner Life}, I put the matter rather forcefully: ``Such philosophically insidious uses of the existential copula are to be banished'' \citep[p.106]{shanahan2010embodiment}. In general, I prefer to ask questions about how words are (or should be) used. The upshot of this is that whenever I say ``LLMs do not literally have beliefs'' (or some such thing), this should be taken as shorthand for ``It is not always appropriate to use the word `belief' (or its relatives) in the context of what an LLM says, even though it would be appropriate if a human being said the same thing'' (or something similar).\footnote{In \citep{shanahan2024simulacra}, I take this approach to consciousness, a far trickier case than belief.}

In short, words like ``literally'', ``really'', and ``just'' should not to be taken as hallmarks of a metaphysical pronouncement, and no sentence in \citep{shanahan2024talking} that uses those words should be taken as endorsing a reductionist view of LLM-based systems. However, this leaves plenty of room for debate over what constitutes appropriate versus inappropriate uses of a word. My paper takes a position on this with respect to belief. The strategy of the paper is to consider a hierarchy of increasingly sophisticated LLM-based systems, noting that, as we ascend the hierarchy, it becomes increasingly appropriate to speak of belief without the need for caveats, exceptions, or clarifications.

At the base of this hierarchy is what I call the ``bare-bones'' LLM. In the strict sense of the term, a ``large language model'' is a function that takes as input a sequence of tokens and returns a probability distribution over tokens representing the model's prediction for the next token in the sequence. This is the bare-bones LLM. It is a computational model of the distribution of words in human language, and it doesn't do anything until it is embedded in a larger system, such as a chatbot app. Confusingly, though, in contemporary usage, the term ``large language model'' or LLM is also used for these larger systems. Hence, people refer to ChatGPT as an LLM, when strictly speaking it is an application built around the core component of an LLM.

While it seems reasonable to allude to the knowledge encoded in a bare-bones LLM, I do think it is misleading to speak of the beliefs of a completely passive entity. To my mind, the very idea of belief is bound up with behaviour. That is to say, the original context for using the word ``belief'' -- its natural home, so to speak -- is living, behaving, active human beings (and other animals), and to use it for a completely passive, inactive, computational entity is to depart too far from the word's original home for comfort. But the bare-bones LLM is hardly an interesting case.

Far more interesting than the bare-bones LLM is the simple LLM-based conversational agent.\footnote{The word here ``agent'' is itself philosophically fraught. By my own lights, I should perhaps be more cautious in its use. But in the field of AI, the term is used in a lightweight technical sense to mean ``anything that can be viewed as perceiving its environment through sensors and acting upon that environment through actuators'' \citep[p.24]{russell2010artificial}, where the environment in question can be a purely textual interface with a human user.} We obtain one of these by embedding the bare-bones LLM in an inner loop that, given the transcript of the conversation so far, repeatedly samples from the distribution output by the model to obtain a sequence of words (the agent's response), and an outer, turn-taking loop that alternates between the user's input and the agent's replies. Now we have a system, based on an LLM, that actually does something, and we can speak of its behaviour. Moreover, we have moved a little closer to the natural home of the word ``belief''. Now suppose the resulting system is very convincing. It is, let us say, a human-level conversationalist. Is it appropriate to use the word ``belief'' in its full sense, without the need for caveats, exceptions, or clarifications? I don't think so. Not in its full sense. Not at this level in the hierarchy of systems we are ascending.

On the one hand, it's perfectly natural to speak loosely of such an agent's beliefs. I might say to a colleague, for example, ``Oh, ChatGPT knows you're a computer scientist, but it thinks you wrote a paper I've never heard of''. In the spirit of Dennett's intentional stance \citep{dennett2009intentional}, this way of talking helps to make sense of the subsequent conversation, and is easier to say than ``ChatGPT's weights predispose it to emit the string XYZ when prompted with the string ABC''. On the other hand, such an agent ``cannot participate fully in the human language game of truth because it does not inhabit the world we human language users share'' \citep[p.73]{shanahan2024talking}. We cannot, for example, ask a simple LLM-based conversational agent whether the pot is full of water. That pot. Look. That one over there. Go and have a look. Is it full or is it empty? The simple LLM-based conversational agent cannot wander over there, peer down, ascertain the status of the pot, and report back to us. Yet this is the sort of primal scene -- a scene wherein a person adjusts what they do and say after engaging with the world and finding something out -- that I see as the original home of the word ``belief'' and its relatives.

However, as we move up the hierarchy of LLM-based systems, layering on more capabilities, the need for caution in using the word ``belief'' gradually lessens. First, we can consider multi-modal LLMs capable of taking visual as well as textual input. Then we can consider LLMs capable of a wider range of actions than merely issuing textual output, a range that could include retrieving web pages or running Python code, for example. Finally, we can consider embodied (or virtually embodied) LLM-based systems, which take input from a camera mounted on a robot or on an avatar in a 3D games-like environment, and whose repertoire of actions includes controlling a robot's effectors or an avatar's movements.

With each of these steps, we move closer to the natural home of the word ``belief'' and its relatives, which is in predicting, explaining, and in general making sense of, the activity of embodied creatures like ourselves, creatures whose behaviour changes in response to what we find out by interacting with the world and the objects it contains. That is the arc of my original paper. The intent was not to take up a metaphysical position with respect to belief, nor to bolster deflationary views of LLM capabilities based on such positions. The aim, rather, was to remind readers of how unlike humans LLM-based systems are, how very differently they operate at a fundamental, mechanistic level, and to urge caution when using anthropomorphic language to talk about them.\footnote{By way of counterpoint, \citep{shanahan2023role} suggests that we can use anthropomorphic language with a degree of impunity if we frame LLM behaviour in terms of role play.}

\bibliography{main}

\end{document}